\documentclass[10pt,twocolumn,letterpaper]{article}

\usepackage{cvpr}
\usepackage{epsfig}
\usepackage{graphicx}
\usepackage{amsmath}
\usepackage{amssymb}

\usepackage[table,xcdraw]{xcolor}
\usepackage{caption}
\usepackage{float}
\usepackage{placeins}
\usepackage{booktabs}
\usepackage{colortbl}
\usepackage{stfloats}
\usepackage{enumitem}
\usepackage[accsupp]{axessibility}

\usepackage[pagebackref,breaklinks,colorlinks]{hyperref}

\usepackage[capitalize]{cleveref}
\crefname{section}{Sec.}{Secs.}
\Crefname{section}{Section}{Sections}
\Crefname{table}{Table}{Tables}
\crefname{table}{Tab.}{Tabs.}

\def\cvprPaperID{4901} 
\def\confName{CVPR}
\def\confYear{2022}

\newcommand{\RNum}[1]{\uppercase\expandafter{\romannumeral #1\relax}}
\newcommand{\coco}{\mbox{\sc{{Coco}}\xspace}}
\newcommand{\rfcoco}{\mbox{\sc{{RefCoco+}}\xspace}}
\newcommand{\vqa}{\mbox{\sc{{VQA v2}}\xspace}}
\newcommand{\cocogpv}{\mbox{\sc{{Coco-sce}}\xspace}}
\newcommand{\model}{\mbox{\sc{{GPV-1}}\xspace}}

\newcommand{\shared}{$\mathcal{S}$}
\newcommand{\hvqacap}{$\mathcal{H}_{vqa,cap}$}
\newcommand{\hclsloc}{$\mathcal{H}_{cls,loc}$}
\newcommand{\best}[1]{\underline{\textbf{#1}}}
\newcommand{\sbest}[1]{\textbf{#1}}
\makeatletter
\newcommand\footnoteref[1]{\protected@xdef\@thefnmark{\ref{#1}}\@footnotemark}
\makeatother

\definecolor{ColorVQA}{HTML}{E6B8AF}
\definecolor{ColorCap}{HTML}{FFF2CC}
\definecolor{ColorLoc}{HTML}{CFE2F3}
\definecolor{ColorCls}{HTML}{D9D2E9}
\definecolor{tabindex}{gray}{0.5}

\usepackage{cleveref}
\begin{document}

\title{Towards General Purpose Vision Systems: \\An End-to-End Task-Agnostic Vision-Language Architecture}

\author{
Tanmay Gupta$^1$ \quad \quad Amita Kamath$^{1}$ \quad \quad Aniruddha  Kembhavi$^1$ \quad \quad Derek Hoiem$^2$\\
$^1$PRIOR @ Allen Institute for AI \quad \quad $^2$University of Illinois at Urbana-Champaign\\
\url{https://prior.allenai.org/projects/gpv}
}

\maketitle



\begin{abstract}
    Computer vision systems today are primarily N-purpose systems, designed and trained for a predefined set of  tasks. Adapting such systems to new tasks is challenging and often requires non-trivial modifications to the network architecture (e.g. adding new output heads) or training process (e.g. adding new losses). To reduce the time and expertise required to develop new applications, we would like to create general purpose vision systems that can learn and perform a range of tasks without any modification to the architecture or learning process. 
    In this paper, we propose  \model, a task-agnostic vision-language architecture that can learn and perform tasks that involve receiving an image and producing text and/or bounding boxes, including classification, localization, visual question answering, captioning, and more. We also propose evaluations of generality of architecture, skill-concept\footnote{For this work, we define concepts, skills and tasks as follows: \textbf{Concepts} -- nouns (\eg \textit{car, person, dog}), \textbf{Skills} -- operations that we wish to perform on the given inputs (\eg classification, object detection, image captioning), \textbf{Tasks} -- predefined combinations of a set of skills performed on a set of concepts (\eg ImageNet classification task involves the skill of image classification across 1000 concepts).} transfer, and learning efficiency that may inform future work on general purpose vision.  Our experiments indicate \model\ is effective at multiple tasks, reuses some concept knowledge across tasks, can perform the Referring Expressions task zero-shot, and further improves upon the zero-shot performance using a few training samples.   
\end{abstract}

\section{Introduction}

Computer vision systems today are $N$-purpose learners --- designed, trained, and limited to $N$ predetermined tasks. Single-purpose models specialize in a single task, and adapting them to a new task or dataset requires an architecture change, minimally replacing the last classification layer. Multi-purpose models, such as Mask-RCNN~\cite{He17Mask}, simultaneously solve more than one task, but the architecture and learning are tailored to specific tasks which must be defined in advance.
In vision-language models~\cite{Lu202012in1MV}, dedicated output heads are typically used for each task and dataset.  


\begin{figure}
    \centering
    \includegraphics[width=0.9\linewidth]{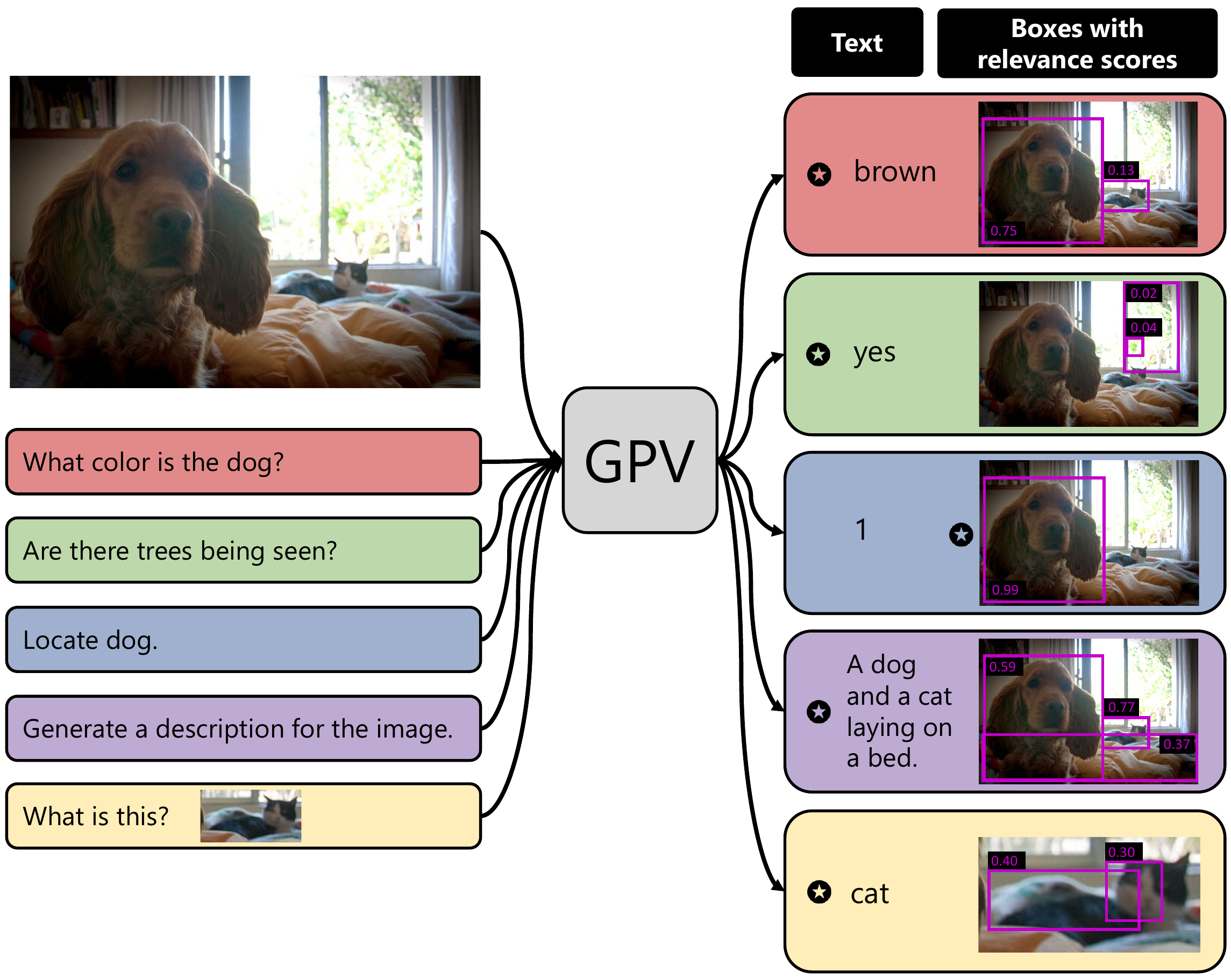}
    \vspace{-0.1in}
    \caption{
    \label{fig:teaser}
    \textbf{A task-agnostic vision-language architecture.} \model\ takes an image and a natural language task description and outputs bounding boxes, confidences and text. \model\ can be trained end-to-end on any task that requires a box or text output, without any architecture modifications such as adding a new task-head. 
    Results correspond to a model trained to perform VQA, localization, captioning, and classification tasks. Star indicates the output modality supervised during training for each task.}
    \vspace{-0.2in}
\end{figure}

Analogous to a general purpose computer, a general purpose vision (GPV) system is designed to carry out many vision tasks, not all known at the time of design, constrained only by its input modalities, memory/instructions, and output modalities. General purpose systems enable new applications to be developed without knowledge of or access to  the underlying mechanics. The NLP community has made significant progress in this direction with sequence-to-sequence transformer-based models, such as T5~\cite{Raffel2020t5} and GPT-3~\cite{Brown2020gpt3}, which can be trained to solve many language tasks without changing the architecture. We believe such advances are now possible within computer vision, though with many new challenges.

In this paper, we propose an end-to-end trainable task-agnostic vision-language architecture, \model, as a step towards general purpose vision systems. 
As input, our system receives an image and a text description of a task. The system outputs bounding boxes, confidences, and text that are relevant to the task and image. A user can input an image and query the system with a variety of requests such as \emph{``What make is the blue car?''} (visual question answering), \emph{``Locate all the sedans''} (localization), and \emph{``Describe the image''} (captioning).  Each query elicits a different response using  output heads that are shared across tasks. Defining the task through natural language allows the user to request \model\ to perform or learn a task without knowledge of its architecture or previous training.  For example, our experiments show that \model\ can perform the referring expressions task without any training examples for that task and, when provided training examples, learns more quickly than special purpose models.  


Beyond performing well on trained skill-concept combinations (contained in training tasks), GPV systems should be able to learn new tasks efficiently with the same architecture as well as generalize to novel skill-concept combinations for learned skills by transferring concept knowledge from other skills. These abilities are not usually applicable or measured in specialized systems. Therefore, we propose evaluations that measure three forms of generality:
\begin{itemize}[leftmargin=*]
\vspace{-0.5em}
\item \textbf{Generality of architecture}: Learn any task within a broad domain specified only through input/output modalities without change to network structure (e.g. learn to classify bird species, without adding new output heads) 
\vspace{-0.5em}
\item \textbf{Generality of concepts across skills}: Perform tasks in skill-concept combinations not seen during training (e.g. localize ``muskrat'' after learning to answer questions about ``muskrats'')
\vspace{-0.5em}
\item \textbf{Generality of learning}: Learn new tasks sample-efficiently with minimal loss to performance on previously learned tasks
\vspace{-0.5em}
\end{itemize}

To test generality of architecture, we train and evaluate our system's ability to perform visual question answering (VQA), captioning, object classification, and object localization on the \coco~dataset~\cite{Lin14Microsoft}, as well as test zero-shot generalization to a referring expression task.  To test generality of concepts across skills, we present a new split of the \coco~images and corresponding task annotations called \cocogpv\ (Skill-Concept Evaluation). In \cocogpv, some concepts (objects) are held-out from each task but exposed via other tasks, 
and then evaluate performance on samples containing held-out concepts. 
To test generality of learning, we fine-tune our system on the referring expressions task and measure its learning curve and extent of forgetting previously learned tasks. 

In summary, our main contributions include: (1) \textbf{An end-to-end trainable, task-agnostic vision-language architecture} for learning and performing classification, grounding, visual question answering, captioning, and other tasks that involve image, text and bounding box modalities. (2) \textbf{Evaluation} that tests generality of architecture, skill-concept transfer, and learning ability.

\section{Related Work}
\textbf{Single-purpose vision-language models.} Over the last decade, specialized and effective approaches have been developed for vision-language tasks, including image captioning~\cite{Farhadi10Every,Karpathy2015Deep,Kulkarni11Baby,Lu18Neural,Vinyals2015Show,Xu2015Show}, phrase grounding~\cite{Plummer15Flickr30K,Plummer17Phrase,Rohrbach16Grounding}, referring expression comprehension~\cite{Kazemzadeh14ReferItGame,Mao16Generation}, visual question answering (VQA)~\cite{Antol15Visual,goyal2017cvpr_balanced_vqa_v2,Wang18FVQA,Hudson19GQA,Wu16Ask,Zellers19VCR},
visual dialog~\cite{Das17Visual}, and text-to-image generation~\cite{Cho2020XLXMERTPC}. Advances that have pushed the performance envelope include cross-model transformer architectures~\cite{Vaswani17Attention}, powerful self-supervised~\cite{Devlin19Bert,Lewis2020BARTDS,Brown2020gpt3} and multitask~\cite{Raffel2020t5} language models, pretrained visual representations from object and attribute detectors~\cite{Anderson2018BottomUpAT,Zhang2021VinVLRV} or text conditioned detectors~\cite{Kamath2021MDETR}, and large-scale image/video-text~\cite{radford2021clip,Jia2021ALIGN,Lei2021ClipBERT,Zellers2021MERLOT} pretraining. 

\textbf{N-purpose vision-language models.} Several recent works aim to unify vision-language tasks with a common architecture. UniT trains a single model for 7 tasks including detection and vision-language tasks but uses task specific heads and does not support captioning. 12-in-1~\cite{Lu202012in1MV} jointly trains VilBERT~\cite{Lu2019Vilbert} on 12 vision-language tasks but with 6 output heads (1 per task group). VL-T5
~\cite{Cho2021VLT5} adapts T5~\cite{Raffel2020t5}, a text-to-text architecture pretrained on a mix of self-supervised and supervised tasks, to jointly train on vision-language tasks with only a text generation head (T5's text decoder). Both of these approaches rely on pre-extracted bounding boxes and region features from an object and attributes detector~\cite{Anderson2018BottomUpAT} and are not end-to-end trainable. E2E-VLP
~\cite{Xu2021E2EVLPEV} presents an end-to-end trainable architecture that is extensively pretrained with masked language modeling, image-text matching, captioning, and object detection objectives, each with a different output head. However, the pretrained model is finetuned separately on each task and therefore does not support multiple tasks with a common set of weights. On the other hand, \model\ is both end-to-end trainable and jointly trained on multiple vision-language tasks. Our architecture takes an image and a textual task description as inputs, and has an output head per modality, namely text, bounding boxes, and relevance scores. Other exciting efforts towards creating general purpose vision architectures include Perceiver
~\cite{Jaegle2021Perceiver} and Perceiver IO
~\cite{Jaegle2021PerceiverIO}, but their potential for multitask learning and utility for vision-language tasks such as VQA and captioning remains to be explored.


\textbf{Task descriptions as a means to architecture generality.} Several works in the natural language domain have tried to blur or erase artificial task boundaries by framing each task as text-to-text transformation with the task specified through a task description. Task descriptions range from templated prompts~\cite{Raffel2020t5} to natural language descriptions~\cite{Mishra2021CrossTaskGV,Weller2020LearningFT}.  
Kumar et al.~\cite{kumar_ask_me_anything_icml16} show that multiple tasks, such as part-of-speech tagging, question answering, and classification, can be formulated as a sequence-to-sequence transformation and solved with a single task-agnostic architecture but with separate parameters trained for each task.  Works such as DecaNLP~\cite{McCann2018decanlp} and UnifiedQA~\cite{Khashabi2020UnifiedQACF} have trained single models to perform multiple tasks by reformulating each task as question answering allowing the individual task-performances to benefit from more data with diverse supervision while sharing model parameters. Works such as T5~\cite{Raffel2020t5}, GPT~\cite{Radford2019gpt2,Brown2020gpt3} have also highlighted the transfer learning capabilities of unified models specially in zero-shot and few-shot scenarios.


\definecolor{darkred}{rgb}{0.56, 0.01, 0.01}
\definecolor{darkblue}{rgb}{0.18, 0.32, 0.56}
\definecolor{darkpurple}{rgb}{0.32, 0.14, 0.47}
\begin{figure*}[h!t]
\centering
\includegraphics[width=0.9\textwidth]{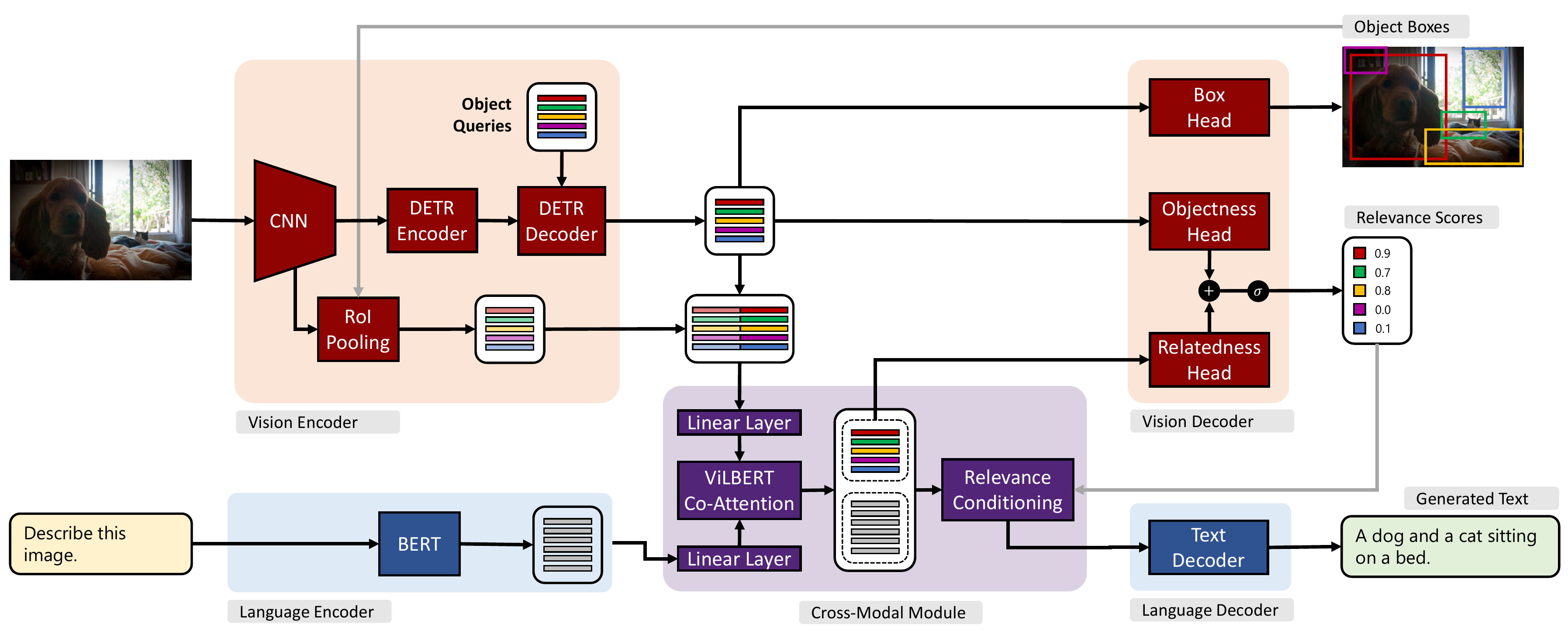}
\vspace{-0.05in}
\caption{
\label{fig:architecture_overview} {\bf Architecture of \model.} \textbf{\textcolor{darkred}{Vision}}, \textbf{\textcolor{darkblue}{language}}, and \textbf{\textcolor{darkpurple}{cross-modal}} modules are color-coded (see Sec.~\ref{sec:vl_architecture} for details). 
}
\vspace{-0.15in}
\end{figure*}

\textbf{Skill-Concept Evaluation.} Few works attempt to learn a concept for one task and apply it to another, e.g. learning to categorize an image as ``aardvark'' and being able to detect or answer questions about aardvarks. 
As one example, \mbox{Gupta et al.}~\cite{Gupta2017AlignedIWR} show that formulating visual recognition and VQA in terms of inner products of word and image-region representations leads to inductive transfer between recognition and VQA tasks. Other works focus on unidirectional transfer to a single task such as captioning~\cite{Hendricks2016DeepCC} or VQA~\cite{Whitehead2021SeparatingSC}. With our \cocogpv\ benchmark, we propose a systematic evaluation of generality of concepts across four standard vision-language tasks by holding out certain concepts from each task while exposing them via other tasks and then measuring performance separately on seen and held-out concepts for each task. 


\section{The \model\ model}
\label{sec:vl_architecture}

\subsection{Architecture Overview}~\label{sec:arch_overview}
The most distinctive aspect of our \model\ system is that tasks are defined through natural language text input, instead of multi-head outputs. Most systems, for example, that perform ImageNet~\cite{Dong09ImageNet} classification and \coco\ detection would have one 1000-class confidence output head and another 80-class box and confidence output head. More tasks or more datasets would require more output heads. 
Once trained, such a system will always produce 1,080 types of confidence and 80 classes of bounding boxes.

\model\ does not have explicit task boundaries and instead takes in a natural language {\em task description} such as ``What is sitting on the sofa?'' (VQA), ``Find all instances of dogs'' (localization), ``What is going on in the image'' (captioning), or ``What kind of object is this?'' (classification).  \model\  interprets and performs all tasks using the same language/vision/cross-modal encoders and decoders. In training, the localization task has bounding box ground truth, while others such as classification, question answering, and captioning have text ground truth. Yet, all tasks involve common skills such as interpreting the task description, localizing objects, representing image regions, and determining relevance to the task.  A new task, such as referring expressions which has bounding box ground truth, can be defined simply by providing new inputs (``Find the man wearing a green shirt'') and output supervision (bounding box). Thus, limited only by modalities that it can sense and produce, \model\ can be trained to perform a wide range of tasks without task-specific modifications to the architecture or learning.



Fig.~\ref{fig:architecture_overview} provides an overview of \model's architecture consisting of a visual encoder, language encoder, vision-language co-attention module, and output heads for the supported output modalities -- boxes, relevance scores, and text. First, we encode the image using the CNN backbone and the transformer encoder-decoder from DETR~\cite{Carion20EndtoEnd}, an end-to-end trainable object detector. Simultaneously, the natural language task description is encoded with BERT~\cite{Devlin19Bert}. Then, to cross-contextualize representations from the visual and language encoders, we use ViLBERT's co-attention module~\cite{Lu2019Vilbert}. Box and objectness heads predict task-agnostic bounding boxes and scores. The relatedness head predicts a task-specific score for each output box that is combined with the objectness scores to obtain relevance scores. The text decoder is an autoregressive transformer decoder that generates text output with relevance-conditioned outputs from cross-modal module serving as memory. 

\subsection{Vision modules}\label{sec:arch_visual}

We use a DETR based \textbf{visual encoder}. A ResNet-50 backbone~\cite{he2016cvpr_resnet} extracts a convolutional feature map that is fed into DETR's transformer encoder to get contextualized features for every grid location. The transformer decoder takes as input $R$ $(=100)$ object queries (learned constant vectors) and the contextualized grid features and produces region descriptors per object query. The main intuition is that the object queries serve as learnable anchors, and the transformer encoder-decoder trained on detection eliminates the need for non-maximum suppression as a post-processing step. The complete region encoding is obtained by concatenating DETR's transformer features, which encode location and limited appearance information, with RoI pooled features  from the CNN backbone.

As a \textbf{vision decoder}, \model\ uses DETR's box head to predict bounding boxes from region descriptors, resulting in $R$ region proposals.  These bounding boxes are used for grounding and detection tasks as well as for RoI pooling from the CNN backbone.
We also replace DETR's 80-way object classification layer with a binary objectness classification layer, which contributes to determining relevance.  

\subsection{Language modules}\label{sec:arch_text}

The \textbf{language encoder} is used to encode the task description. We use BERT's WordPiece tokenizer~\cite{wu2016arxiv_wordpiece} to obtain sub-word tokens for the language input and a pre-trained BERT model to compute representations. Sub-word tokenization provides robustness to out-of-vocabulary words, and large scale language model pretraining allows \model\ to better handle paraphrases of language queries and zero-shot generalization to novel task descriptions, assuming semantic similarity to previously seen descriptions in the BERT embedding space.    

The \textbf{language decoder} outputs words to classify, describe, or answer the input.  Specifically, the sequence of co-attended region representations and language query's token representations are concatenated to construct a single sequence that serves as memory for the transformer text decoder. At each generation step, the sequence of words generated thus far are fed into the decoder along with the memory and a distribution over the vocabulary words is predicted to sample the next word. The inputs to the transformer decoder are trainable word embeddings. The output logit for a vocabulary word is obtained by taking dot product between the embedding vector output by the decoder and a linearly transformed BERT encoding of the word. 

\subsection{Cross-modal modules}

The region descriptors from the vision modules and sub-token representations from the language module are transformed by linear layers to equal dimension vectors and fed into ViLBERT's \textbf{co-attention} layers for cross-contextualization. The relatedness head uses the co-attended region features to predict logits that indicate relevance of regions to the task description. These logits are added to logits from the objectness head and transformed into region-relevance scores by a sigmoid activation. These relevance scores are used to rank bounding boxes or indicate importance of regions to performing the task.


\textbf{Relevance conditioning} modulates the co-attended visual features with relevance scores. Specifically, the relevance score $s$ of each region is used to weight learned vectors $\{v_{\text{rel}},v_{\text{nrel}}\}$, which are added to the region features before feeding to the decoder. This conditioning enables supervision from the text decoder to affect the relatedness and objectness heads. In this way, a model trained to produce captions for images of peacocks may learn to localize peacocks, and, conversely, the ability to localize peacocks may translate to improved caption quality.

\subsection{Training}
\label{sec:pretraining}

Each training sample consists of an image, a task description, and targets. Depending on the task, targets could consist of ground truth bounding boxes, text, or both.  In each training iteration, we uniformly draw samples across all tasks to construct mini-batches. For all samples that contain a text target, we maximize the log-likelihood of the ground truth text. For all samples that contain bounding boxes as targets, we use DETR's Hungarian loss for training the box and relevance prediction.

\noindent\textbf{Initialization.} We initialize all vision modules except the last linear layer in the objectness head with weights from DETR pretrained on either \coco\ or \cocogpv\ (Sec.~\ref{sec:data}) object detection data. 
BERT is pretrained on BooksCorpus~\cite{zhu_bookcorpus_2015iccv} and English Wikipedia.

\noindent\textbf{Optimization.} We train \model\ with a batch size of $120$ and AdamW optimizer~\cite{loshchilov2018decoupled_adamw}. We keep DETR weights frozen for the first $10$ epochs and finetune all modules except BERT for $30$ more epochs. For learning rate (LR), we do a warm-up over the first $4$ epochs to a maximum of $10^{-4}$ followed by linear decay to $0$. Following DETR, we apply gradient clipping on visual module parameters and use a maximum learning rate of $10^{-5}$ for the CNN backbone. We use a $0.05\times$ lower text loss weight for captioning since more words are in the target text than other tasks. 

\section{Tasks and Data}

Our experiments involve 5 tasks using images from the \coco\ dataset and annotations from the \coco, \vqa~\cite{goyal2017cvpr_balanced_vqa_v2}, and \rfcoco~\cite{Kazemzadeh14ReferItGame} datasets. Sec.~\ref{sec:tasks} describes how these tasks are posed to our general purpose system along with respective losses and metrics used for training and evaluation. Sec.~\ref{sec:data} details how samples are created for each task from the original annotations and introduces our \cocogpv\ split for testing the generalization of concepts across skills.

\subsection{Tasks}\label{sec:tasks}

Our experiments mainly involve 4 tasks -- VQA, Captioning, Localization, and Classification. We only use Referring Expressions to test the learning ability of \model. 

\noindent\textbf{VQA} aims to answer a question given an image. The input is an image/text pair, and the output is text. While training, the loss employed is the negative log likelihood of the ground truth answer text. We use the standard VQA evaluation metric (annotator-agreement weighted answer accuracy)~\cite{Antol15Visual} to report results.
    
\noindent\textbf{Captioning} aims to produce a description of an image.  The input is an image and a prompt, such as ``Describe the image'' or ``What is going on in the image?'', and the output is text.  While training, the loss employed is the negative log likelihood of the annotated caption. The evaluation metric reported is CIDEr-D~\cite{Vedantam2015CIDErCI} that measures the similarity of the generated and ground truth captions.
    
\noindent\textbf{Localization} aims to produce a tightly fitting bounding box to an object. The input is an image and a prompt, such as ``Find all instances of dogs'' or ``Locate the chairs'', and the output is a set of ranked bounding boxes.  Training uses DETR's Hungarian loss. Evaluation is an average of per-query average-precision (AP) with a 0.5 bounding box intersection over union (IOU) threshold.  For example, if an image contains two target objects and the correctness of the top four ranked boxes is \{True, False, False, True\}, the AP is (1/1+2/4)/2=0.75 (every-point interpolation). The reported number is AP averaged over samples.

\noindent\textbf{Classification} aims to assign a category to a region.  The input is an image patch and a prompt such as ``What is this thing?'' or ``What object is this?'', and the output is text. In principle, \model\ can produce any category label within the large vocabulary of the text decoder, including words that it has not seen within its classification training data. However, for evaluation, a K-way classification is performed by suppressing outputs that do not correspond to any of the applicable K categories. The training loss used is the negative log likelihood of text output, and evaluation is accuracy averaged over samples.

\noindent\textbf{Referring expressions (RefExp)} aims to localize a single region that corresponds to a phrase.  The input is an image and a referring expression such as ``the man wearing a green shirt'', and the output is one bounding box.  While the training loss and evaluation is the same as localization, the key distinction is disambiguation of the referred instance among other instances of the same object category in the image. 

\subsection{Data}\label{sec:data}

We present experiments using images from the richly annotated \coco\ dataset. We use question and answer annotations from the \vqa\ dataset, referring expressions from \rfcoco, and \coco\ annotations for other tasks.

\noindent\textbf{Data samples.} VQA samples consist of the original questions as prompts paired with the most-agreed answer among annotators. For captioning, \coco\ provides $5$ captions per image, each of which is treated as a different sample paired with one of 14 captioning prompt templates. We generate localization samples for each object category in the image using one of 18 prompt templates paired with all instances for the category. For classification, we create a sample for each object category in the image by choosing one of the instances (cropped using the ground truth box) paired with one of 4 prompt templates. RefExp samples consist of referring expressions as prompts with corresponding boxes.

\noindent\textbf{Data splits.} We present results for \model\ and baselines on two data splits. First we train and evaluate models using the standard data splits for the corresponding tasks. This provides results for \model\ in the context of past work. Then, to test the ability of vision systems to generalize concepts across skills, we present a new split of the above annotations, named \cocogpv\ (Skill-Concept Evaluation).

\begin{figure}[t]
\centering
\includegraphics[width=0.85\linewidth]{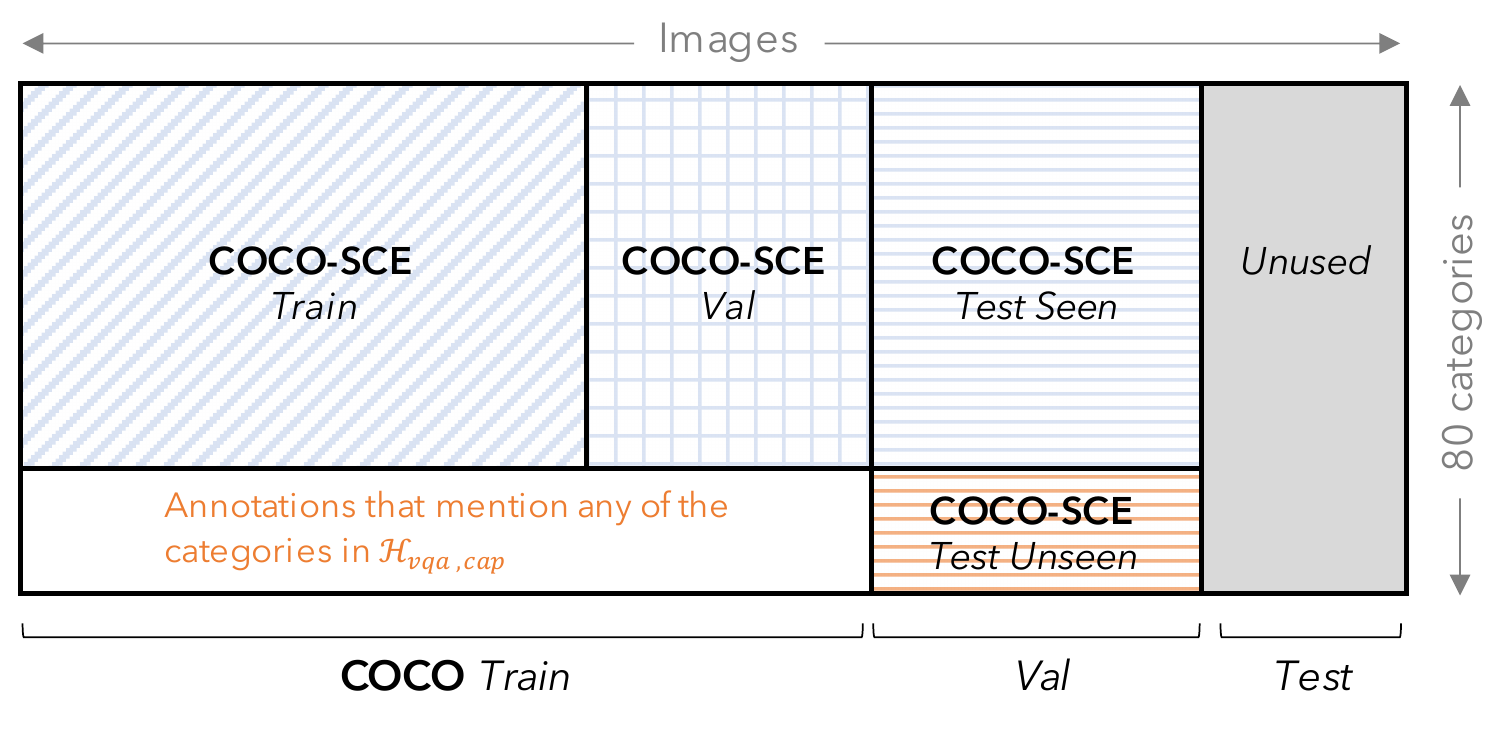}
\vspace{-0.13in}
\caption{
\label{fig:coco-sce} \cocogpv : A split of \coco\ images and annotations to test the generalization of concepts across skills. Schematic shows train, val and test samples used for VQA.
}
\vspace{-0.2in}
\end{figure}

\noindent\textbf{COCO-SCE.} Fig.~\ref{fig:coco-sce} presents a schematic of the proposed \cocogpv\ splits. The 80 classes of \coco\ are split into 3 disjoint sets, specifying which tasks can use them for training and validation:
\begin{itemize}[leftmargin=*]
\vspace{-0.5em}
    \item \hvqacap: 10 classes held-out from the VQA and captioning tasks in the  train/val sets 
    \item \vspace{-0.8em} \hclsloc: 10 different classes held-out from the classification and localization tasks in the train/val sets 
    \item \vspace{-0.8em} \shared: 60 remaining classes are not held out from any tasks
\vspace{-0.5em}
\end{itemize}

When a category is held out, any annotations containing that word are not used for training or val.  E.g., if {\em boat} is a held out category for VQA, then the annotation \{``What color is the boat?'', ``Blue''\} would be excluded from the train/val set. Other annotations from the same image may still be used, e.g. \{``Is it a sunny day?'', ``Yes''\}.   Also, the classification and localization annotations for {\em boat} would be included in train/val for respective tasks.
The assignment of categories to \hvqacap, \hclsloc, and \shared\ is random, except that we assign {\em person} to \shared, because it is so common. 

Images in \cocogpv\ train and val sets come from \coco\ train set, and images in \cocogpv\ test set are those in the \coco\ validation set (as \coco\ test annotations are hidden). 
\cocogpv\ train and val splits are created by first creating an 80-20 partition of \coco\ train images and then for each task discarding samples that expose the held-out categories for that task through the annotations. 
On the test set we report performance separately for samples belonging to ``seen" (e.g. \mbox{\shared\ $\cup$\ \hclsloc} for VQA)  and ``unseen" (e.g. \hvqacap\ for VQA) categories for each task. 


\section{Experiments}
Our experiments evaluate \model\ for its effectiveness compared to specialized models (Sec.~\ref{sec:effectiveness}), its ability to apply learned skills to unseen concepts for that skill (Sec.~\ref{sec:skill-concept-gen}), its efficiency at learning new skills, and retention of previously learned skills (Sec.~\ref{sec:exp_learning}). Sec~\ref{sec:ablations} provides ablations. Our \cocogpv\ experiments are carefully designed to ensure that compared methods train on the same amount of skill data (although some models may have access to data from another skill) and to 
enable evaluation of concept transfer across skills by avoiding exposing the held-out concepts via pretraining on Conceptual Captions~\cite{Sharma18Conceptual} or Visual Genome~\cite{Anderson2018BottomUpAT}. ImageNet pretraining while not ideal, is unavoidable as most vision models including DETR rely on it to bootstrap learning. 



\begin{table}[]
\setlength{\tabcolsep}{3pt}
\small
\centering
\begin{tabular}{ll>{\columncolor{ColorVQA}}c>{\columncolor{ColorCap}}c>{\columncolor{ColorLoc}}c>{\columncolor{ColorCls}}c}
\toprule
Split & Model & VQA & Cap. & Loc. & Class. \\
\midrule
\cocogpv\ & \textcolor{tabindex}{[a]} Specialized Model   & 56.6              & 0.832             & 62.4         & 75.2 \\
& \textcolor{tabindex}{[b]} 1-Task \model\     & 55.9     &0.855    & \textbf{64.8}                  & 75.3 \\
& \textcolor{tabindex}{[c]} Multitask \model\     & \textbf{58.8}     & \textbf{0.908}    & 64.7                  & \textbf{75.4} \\
\midrule
\coco\ & \textcolor{tabindex}{[d]} Specialized Model   & 60.1              & 0.961             & \textbf{75.2}         & 83.3 \\
& \textcolor{tabindex}{[e]} Multitask \model\     & \textbf{62.5}     & \textbf{1.023}    & 73.0                  & \textbf{83.6} \\
\bottomrule                                    
\end{tabular}
\vspace{-0.05in}
\caption{
\label{tab:Generality_vs_Effectiveness}
\textbf{Comparison to special purpose baselines (\cocogpv\ and \coco\ splits)}: Our jointly trained \model\ compares well to specialized single-task baselines as well as \model\ trained on individual task data. On \coco\ split, we report test-server results for VQA and captioning and validation results for localization and classification as the annotations for test images are hidden. On \cocogpv\ split, we report test results for all tasks.}
\vspace{-0.15in}
\end{table}

\begin{table*}
\setlength{\tabcolsep}{5pt}
\small
\centering
\begin{tabular}{l>{\columncolor{ColorVQA}}c>{\columncolor{ColorVQA}}c>{\columncolor{ColorVQA}}c>{\columncolor{ColorCap}}c>{\columncolor{ColorCap}}c>{\columncolor{ColorCap}}c>{\columncolor{ColorLoc}}c>{\columncolor{ColorLoc}}c>{\columncolor{ColorLoc}}c>{\columncolor{ColorCls}}c>{\columncolor{ColorCls}}c>{\columncolor{ColorCls}}c}

\toprule
\multicolumn{1}{l}{} & \multicolumn{3}{c}{\cellcolor[HTML]{E6B8AF}VQA}                                       & \multicolumn{3}{c}{\cellcolor[HTML]{FFF2CC}Captioning}                                               & \multicolumn{3}{c}{\cellcolor[HTML]{CFE2F3}Localization}                          & \multicolumn{3}{c}{\cellcolor[HTML]{D9D2E9}Classification}                           \\

\multicolumn{1}{c}{Model} & \cellcolor[HTML]{E6B8AF}Test        & \cellcolor[HTML]{E6B8AF}\emph{Seen} & \cellcolor[HTML]{E6B8AF}\emph{Unseen} & \cellcolor[HTML]{FFF2CC}Test & \cellcolor[HTML]{FFF2CC}\emph{Seen} & \cellcolor[HTML]{FFF2CC}\emph{Unseen} & \cellcolor[HTML]{CFE2F3}Test & \cellcolor[HTML]{CFE2F3}\emph{Seen} & \cellcolor[HTML]{CFE2F3}\emph{Unseen} & \cellcolor[HTML]{D9D2E9}Test & \cellcolor[HTML]{D9D2E9}\emph{Seen} & \cellcolor[HTML]{D9D2E9}\emph{Unseen} \\
\midrule
\textcolor{tabindex}{[a]} Specialized Model           & 56.6                               & 57.2                         & 45.2                                    & 0.832  & 0.867                        & 0.501                                   & 62.4               & 68.1                & 7.4                        & 75.2                        & 83.0                & 0.0                            \\
\textcolor{tabindex}{[b]} 1-Task \model\           & 55.9                               & 56.5                         & 41.9                                    & 0.855  & 0.891                        & 0.524                                   & \textbf{64.8}               & \textbf{69.8}                & 16.4                        & 75.3                        & \textbf{83.1}                & 0.0                            \\
\textcolor{tabindex}{[c]} Multitask \model\        & \textbf{58.8} & \textbf{59.3}                & \textbf{47.7}                           & \textbf{0.908}              & \textbf{0.944}               & \textbf{0.560}                          & 64.7                        & 68.8                         & \textbf{25.0}               & \textbf{75.4}               & 82.6                         & \textbf{5.4}                   \\
\midrule
\textcolor{tabindex}{[d]} Multitask \model\ \textbf{Oracle} & 61.4                               & 61.3                         & 64.0                                    & 1.018                       & 0.997                        & 0.939                                   & 73.0                        & 72.7                         & 76.0                        & 83.6                        & 83.4                         & 85.7           \\
\bottomrule
\end{tabular}
\vspace{-0.1in}
\caption{
\label{tab:concept_skill_generalization}
\textbf{Skill-Concept Generalization}: Multitask achieves higher performance overall, especially for ``Unseen'' concepts. Classification and Localization ``Seen'' performance slightly decreases, likely because all tasks share the same images and VQA and captioning are more weakly supervised. GPV oracle performance, with no concepts held out, provides an upper-bound on ``Unseen''.	Rows \emph{a,b,c} are trained and tested on the smaller \cocogpv\ data split, while \emph{d} uses the \coco\ split.			
}
\vspace{-0.1in}
\end{table*}

\subsection{Generality \vs Effectiveness}\label{sec:effectiveness}

Is generality of \model\ at the cost of effectiveness? We compare \model\ to competitive special purpose models designed for each task -- ViLBERT~\cite{Lu2019Vilbert} (VQA), VLP~\cite{zhou2019_vlp} (captioning), Faster-RCNN~\cite{ren_2015_fasterrcnn}) (localization) and Resnet-50~\cite{he2016cvpr_resnet} (classification). To avoid conflating effectiveness of architecture with availability of more data, we retrain these models to only use \coco\ and \vqa\ annotations. For ViLBERT and VLP this requires replacing Visual Genome~\cite{Krishna17Visual} bottom-up features~\cite{Anderson2018BottomUpAT} with Faster-RCNN features trained only on \coco\ and no pretraining on Conceptual Captions~\cite{Sharma18Conceptual}.

Tab.~\ref{tab:Generality_vs_Effectiveness} shows that on the \cocogpv\ split, the general purpose \model\ architecture trained on individual tasks compares favorably to each special purpose model (rows \emph{a} vs \emph{b}). Also, the generality of \model\ enables it to be jointly trained on all 4 tasks, leading to sizeable gains on 2 tasks and comparable results on others (rows \emph{b} vs \emph{c}). The same trends also hold when we compare models on the original \coco\ data-splits (rows \emph{d} vs \emph{e}), validating that these trends are not merely a product of our proposed splits. Together, these results establish that the generality of \model\ is not at the expense of effectiveness.

\subsection{Skill-Concept Generalization}\label{sec:skill-concept-gen}
We wish to test generality of concepts across skills, \ie the ability of a model to perform well on novel skill-concept combinations that were unseen during training. When training on a single task on \cocogpv\, a model does not have access to any annotation on held-out concepts. For example, a model trained only on VQA will never see a question or answer about {\em horse} $\in$ \hvqacap. However, when training on all tasks, the model learns to localize and classify {\em horse} images. Therefore we expect the model to apply the acquired skill of question answering to answer questions about {\em horse} without explicitly being trained on {\em horse} VQA data.

Tab.~\ref{tab:concept_skill_generalization} shows the performance of the specialized models and the 1-Task and Multitask \model\ models on the \cocogpv\ full test split as well as separately on the subset of test data categorized as ``seen" and ``unseen" (see Fig.~\ref{fig:coco-sce} for a schematic of these subsets for the VQA task).  The 1-Task \model\ (row \emph{b}) trained on individual tasks serves as a baseline to account for learned priors and dataset biases by the \model\ architecture. We observe significant gains by Multitask \model\ (row \emph{c}) on the ``unseen" subset across all tasks, particularly over the specialized models (row \emph{c} vs row \emph{a}) -- indicating that the general purpose architecture is better suited at learning skills and then applying them to concepts that were unseen for that skill. We also report the performance of Multitask \model\ trained on the \coco\ training split (row \emph{d}). Since this split exposes the model to held-out concepts for all tasks, it can serve as a loose upper bound for the ``unseen" split. 
\subsection{Learning Generalization}
\label{sec:exp_learning}

A system exhibits good learning generalization if it can learn new skills sample-efficiently without forgetting previously-learned skills. 

\begin{figure}[t]
\centering
\includegraphics[width=0.95\linewidth]{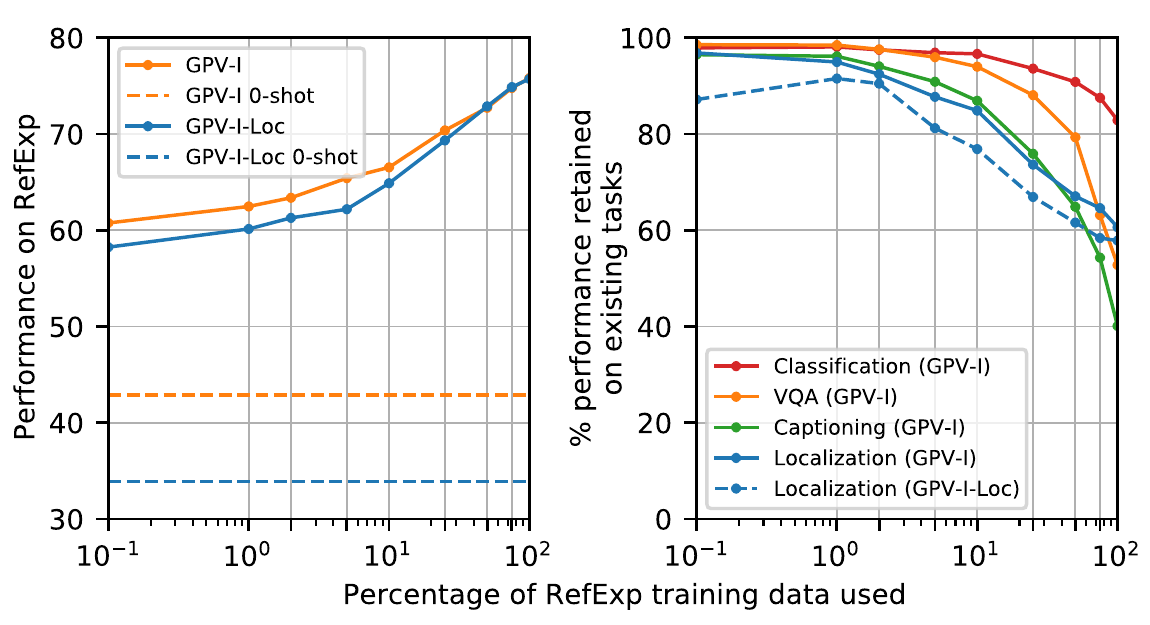}
\vspace{-0.1in}
\caption{
\label{fig:refexp} {\bf Learning new skills and retention of previous skills.} \emph{Left:} On \rfcoco, multitask pretrained \model\ improves the 0-shot performance over single task pretrained \model-Loc. \model\ also learns the new skill quicker than \model-Loc, particularly in the lower data regime. \emph{Right:} As \rfcoco\ training data increases, \model\ does forget existing skills but multitask \model\ is more resilient to forgetting than \model-Loc (note that $x$-axes are log-scaled).
}
\vspace{-0.15in}
\end{figure}

\noindent
\textbf{Learning ability.} Fig.~\ref{fig:refexp} (left) shows learning curves for \model\ and \model-Loc when finetuning on the Referring Expressions task. \model-Loc is pretrained on only the localization task (the only other task that has bounding-box supervision) while \model\ is pretrained on all four tasks.  Multitask \model\ demonstrates much better zero-shot performance as well as better sample-efficiency in the low data regime. The learning of attributes and additional nouns provides a better starting point for referring expressions; e.g., while the localization-trained model starts with the ability to localize {\em person}, the multitask model is also familiar with {\em red} and {\em sweater} through captions and VQA and may better localize ``the person wearing a red sweater''.



\noindent
\textbf{Retention.} Fig.~\ref{fig:refexp} (right) shows the percent of performance retained on the original tasks as \model\ is trained with increasing amounts of \rfcoco\ training data. 
Interestingly, Multitask \model\ forgets slower than \model-Loc on the localization task. Localization and captioning suffer the most from catastrophic forgetting while classification shows robust retention. \model\ does not have explicit mechanisms for addressing forgetting, but our results highlight the importance of such mechanisms for general purpose learning. 

\begin{table}[t]
\setlength{\tabcolsep}{5pt}
\small
\centering
\begin{tabular}{l>{\columncolor{ColorVQA}}c>{\columncolor{ColorCap}}c>{\columncolor{ColorLoc}}c>{\columncolor{ColorCls}}c}
\toprule
 & VQA & Cap. & Loc. & Class. \\
\midrule
\textcolor{tabindex}{[a]} Multitask \model\        & \textbf{58.8}                        & \textbf{0.908}                        & 64.7                                  & 75.4                                    \\
\textcolor{tabindex}{[b]} \hspace{0.4cm} \emph{w/o RoI features}     & 54.9                                 & 0.898                                 & \textbf{65.3}                         & \textbf{76.6}                           \\
\textcolor{tabindex}{[c]} \hspace{0.4cm} \emph{w/o Fine-Tuning}      & 56.4                                 & 0.883                                 & 63.4                                  & 71.5                  \\
\bottomrule
\end{tabular}
\vspace{-0.05in}
\caption{
\label{tab:ablation}
\textbf{Ablation}: 
Augmenting the vision transformer features with RoI features extracted from the CNN backbone helps VQA significantly and Captioning slightly, but is detrimental to Localization and Classification. The transformer features may be sufficient to fully model localization and classification, while VQA and Captioning benefit from additional information in the RoI features. Fine-tuning helps all tasks.
}
\vspace{-0.2in}
\end{table}

\subsection{Ablations}\label{sec:ablations}

Tab.~\ref{tab:ablation} ablates key factors that make \model\ effective. Finetuning end-to-end (as opposed to keeping DETR weights frozen) contributes to performance across all tasks (rows \emph{a} vs \emph{c}). RoI pooling significantly boosts performance for VQA, slightly for captioning, but leads to slight drop for localization and classification (rows \emph{a} vs \emph{b}).

\begin{figure*}[h!]
\centering
\includegraphics[width=0.99\textwidth]{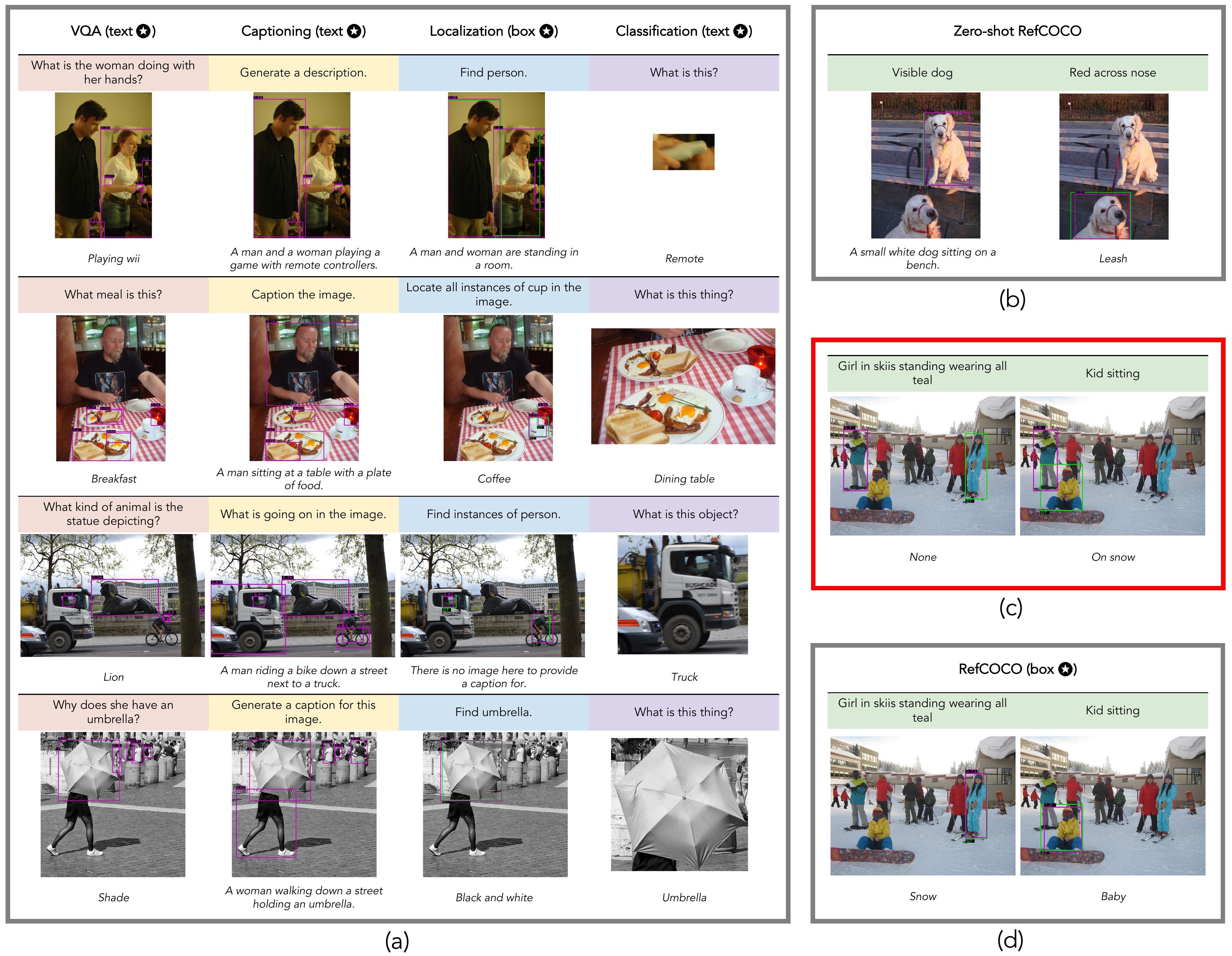}
\vspace{-0.5em}
\caption{
\label{fig:qualitative_analysis} {\bf Qualitative Results:} Prompts are shown in colored boxes (one color per task) with box and text predictions below. (a) \model\ learns to output the expected modality (indicated by star) for each task, but also provides unsolicited yet informative commentary for the localization task and relevant regions for VQA and captioning. (b) \model\ can perform 0-shot referring expression comprehension. \model\ learns to correct zero-shot mistakes (c) when finetuned on annotations for the task (d). 
}
\end{figure*}



\begin{figure*}[h!]
\centering
\includegraphics[width=0.99\textwidth]{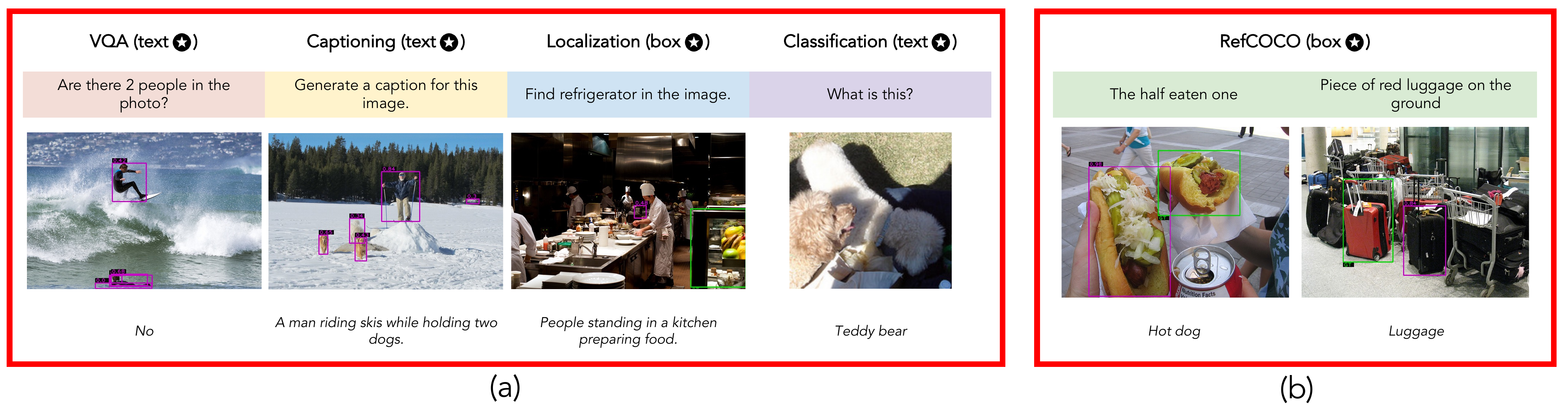}
\vspace{-0.5em}
\caption{
\label{fig:negative_qualitative} {\bf Failure Cases:} (a) \model\ fails to count the number of people despite localizing them, and is unable to locate objects such as ski poles and refrigerators. (b) \rfcoco\ failures showing the model locating an incorrect object from the correct category.
}
\end{figure*}




\section{Limitations and Conclusion}

The \model\ architecture can be trained to perform any image task that can be described and performed using words or boxes.  Our experiments show that this generality does not come at the expense of accuracy, as \model\ compares well to specialized systems when trained on individual tasks and outperforms when trained jointly. However, several challenges remain. Generality of \model\ is at the cost of runtime efficiency compared to specialized systems. For example, using \model\ for detection requires a separate localization inference per object category. \model\ also achieves some skill-concept generalization, as measured on our \cocogpv\ split, but a large gap to oracle indicates significant room for improvement. 
Our referring expression comprehension experiments show that while \model\ learns more quickly and forgets more slowly when trained on multiple tasks, catastrophic forgetting remains a challenge. While \cocogpv\ does provide a controlled test bed for studying GPVs, our evaluation is limited to skills and concepts based on \coco. 
Finally, due to lack of an image generation head, \model\ currently does not support image manipulation or generation tasks such as colorization and segmentation. \model\ also does not handle non-image inputs such as videos or point clouds. Extending the capabilities of \model\ to new tasks and input and output types is an exciting challenge for future work. 



\vspace{0.05in}
\noindent\textbf{Supplemental material} contains additional training and dataset details, task prompts, ablations, analysis, potential negative impacts, and qualitiative results.

\vspace{0.05in}
\noindent\textbf{Acknowledgments.} This work is partially supported by ONR MURI Award N00014-16-1-2007 and ONR award N00014-21-1-2705.

{\small
\bibliographystyle{ieee_fullname}
\bibliography{derek_bib}
}

\appendix
\section{Summary}
This supplementary material contains:
\begin{itemize}
    \vspace{-0.5em} \item Training details (Sec.~\ref{sec:train_details})
    \vspace{-0.5em} \item Additional ablation results (Sec.~\ref{sec:additional_ablation})
    \vspace{-0.5em} \item Task descriptions (Sec.~\ref{sec:task_descr})
    \vspace{-0.5em} \item Dataset statistics (Sec.~\ref{sec:stats})
    \vspace{-0.5em} \item Potential Negative Impact (Sec.~\ref{sec:neg_impact})
    \vspace{-0.5em} \item Randomly sampled \href{https://prior.allenai.org/projects/gpv#qualitative-results}{qualitative results}
    \vspace{-0.5em} \item \href{https://github.com/allenai/gpv-1}{Code} available under Apache-2.0 license.
\end{itemize}

\section{Training Details}\label{sec:train_details}
\textbf{Mini-batch sampling strategy.} During training, batch sizes are created by randomly drawing samples uniformly across all tasks. Alternatives that require further exploration include drawing samples from only one task in each iteration or ensuring an equal representation of samples from each task in a mini-batch. 

\textbf{Learning from box supervision.} For samples that contain a ground truth bounding box (localization task), we compute DETR's Hungarian loss. The Hungarian loss first pairs ground truth and predicted boxes. Hungarian algorithm is used for matching using a linear combination of three cost functions that evaluate for label correctness (relevant or background), high overlap (computed as generalized IoU), and low L1 distance between true and predicted box coordinates. For each pair, the Hungarian loss then minimizes the negative log-likelihood of the ground truth class, generalized IoU loss, and L1 coordinate regression loss.

\textbf{Learning from text supervision.} For VQA, captioning, and classification tasks which have a text target, we minimize the negative log-likelihood of the ground truth text. This is implemented using cross-entropy with one-hot targets for each word in the text. Ideally, this could be implemented by gathering all samples with text supervision in the mini-batch and minimizing the mean negative log-likelihood (mean computed over samples and not words). However, this led to early overfitting for captioning while other task performances were still improving on the respective validation sets. This is due to a large difference in the number of words in the target text in captioning (upto $20$ words) vs other tasks ($1$-$5$ words). Therefore we use a weight of $0.05$ ($=1/20$) for captioning samples while using a weight of 1 for other samples which addressed the early overfitting issue.  

\textbf{Reproducibility.} Pretrained models and training scripts are available at \url{https://github.com/allenai/gpv-1} under Apache-2.0 license.

\textbf{Hardware Requirements.} We have trained \model\ models using either 4$\times$ Titan RTX (24GB), 4$\times$ Quadro RTX 8000 (48GB), or 1$\times$ A100 Tensor Core GPU (80GB). On the single A100, \model\ requires slightly under 60GB of memory but more on the other setups due to multi-GPU training overhead. We have been unable to finetune the full model with 120 batch-size on 8$\times$ 12 or 16GB GPUs due to memory issues but it may be possible to further optimize the implementation to do so (e.g using mixed precision training). With batch-size of 120, training \model\ on all 4 tasks takes 5-7 days depending on the hardware and training split. It may be possible to trade off memory requirements with training time by reducing the batch-size but may require hyper-parameter tuning to retain performance.

\textbf{Efficiency metrics.} \model\ has 236M parameters, and for a 640x480 image (the image size used during training) on a GeForce RTX 2080 Ti, the inference forward pass yields ~289M activations and ~139G flops depending on the length of the task description and output text.

\section{Additional Ablation Results}\label{sec:additional_ablation}
In Tab.~\ref{tab:additional_ablation} we show a more detailed task ablation. In addition to performance of \model\ trained on individual tasks and all tasks, we also provide results for \model\ trained on VQA and captioning which does not see any concept in \hvqacap. Hence, we do not expect significant gains on VQA and captioning unseen sets. However, \mbox{\model-VQA+Cap+Loc}
introduces \hvqacap\ categories during training through the localization task allowing the model to improve the performance on VQA and captioning unseen sets. \model-Multitask results in additional gains by letting the model benefit from classification supervision as well. Similarly, \model-Cls+Loc continues to get zero unseen classification accuracy as \hclsloc\ concepts are held out during training but introducing those concepts through VQA and captioning in \model-Multitask leads to improved unseen classification accuracy.  

\begin{table*}
\setlength{\tabcolsep}{5pt}
\small
\centering

\resizebox{\textwidth}{!}{%
\begin{tabular}{lc>{\columncolor{ColorVQA}}c>{\columncolor{ColorVQA}}c>{\columncolor{ColorVQA}}c>{\columncolor{ColorCap}}c>{\columncolor{ColorCap}}c>{\columncolor{ColorCap}}c>{\columncolor{ColorLoc}}c>{\columncolor{ColorLoc}}c>{\columncolor{ColorLoc}}c>{\columncolor{ColorCls}}c>{\columncolor{ColorCls}}c>{\columncolor{ColorCls}}c}
\toprule
                  & \multicolumn{1}{c}{*Concepts seen during training} & \multicolumn{3}{c}{\cellcolor[HTML]{E6B8AF}VQA} & \multicolumn{3}{c}{\cellcolor[HTML]{FFF2CC}Captioning} & \multicolumn{3}{c}{\cellcolor[HTML]{CFE2F3}Localization} & \multicolumn{3}{c}{\cellcolor[HTML]{D9D2E9}Classification}                                           \\
         \multicolumn{1}{c}{Model}         & \shared \hspace*{0.5cm} \hvqacap \hspace*{0.4cm} \hclsloc & Test & \emph{Seen} & \begin{tabular}[c]{@{}c@{}}\emph{Unseen}\\(\hvqacap)\end{tabular} & Test  & \emph{Seen} & \begin{tabular}[c]{@{}c@{}}\emph{Unseen}\\(\hvqacap)\end{tabular} & Test &  \emph{Seen} & \begin{tabular}[c]{@{}c@{}}\emph{Unseen}\\(\hclsloc)\end{tabular} & Test &  \emph{Seen} & \begin{tabular}[c]{@{}c@{}}\emph{Unseen}\\(\hclsloc)\end{tabular}\\ 
         \midrule
\textcolor{tabindex}{[a]} \model-VQA & \checkmark \hspace*{2.7cm} \checkmark \hspace*{0.3cm} & 55.9 & 56.5 & 41.9 & - & - & - & - & - & - & - & - & - \\
\textcolor{tabindex}{[b]} \model-Cap & \checkmark \hspace*{2.7cm} \checkmark \hspace*{0.3cm} & - & - & - & 0.855 & 0.891 & 0.524 & - & - & - & - & - & - \\
\textcolor{tabindex}{[c]} \model-Loc & \checkmark \hspace*{1cm} \checkmark \hspace*{2cm} & - & - & - & - & - & - & 64.8 & 69.8 & 16.4 & - & - & - \\
\textcolor{tabindex}{[d]} \model-Cls & \checkmark \hspace*{1cm} \checkmark \hspace*{2cm} & - & - & - & - & - & - & - & - & - & \sbest{75.3} & \best{83.1} & 0.0 \\ 
\midrule
\textcolor{tabindex}{[e]} \model-VQA+Cap & \checkmark \hspace*{2.7cm} \checkmark \hspace*{0.3cm} & 57.6 & 58.3 & 42.7 & 0.876 & 0.913  & 0.536 & - & - & - & - & - & - \\
\textcolor{tabindex}{[f]} \model-Cls+Loc & \checkmark \hspace*{1cm} \checkmark \hspace*{2cm} & - & - & - & - & - & - & \sbest{64.9} & \best{70.0} & 16.3 & 74.5 & 82.2 & 0.0\\
\textcolor{tabindex}{[g]} \model-VQA+Cap+Loc & \checkmark \hspace*{1cm} \checkmark \hspace*{1.35cm} \checkmark \hspace*{0.3cm} & \best{59.6} & \best{60.2} & \sbest{46.6} & \best{0.911} & \best{0.949} & \sbest{0.559} & \best{65.1} & \sbest{69.3} & \sbest{24.1} & - & - & - \\ 
\midrule
\textcolor{tabindex}{[h]} \model-Multitask & \checkmark \hspace*{1cm} \checkmark \hspace*{1.35cm} \checkmark \hspace*{0.3cm} & \sbest{58.8} & \sbest{59.3} & \best{47.7} & \sbest{0.908} & \sbest{0.944} & \best{0.560} & 64.7 & 68.8 & \best{25.0} & \best{75.4} & \sbest{82.6} & \best{5.4} \\
\bottomrule         
\end{tabular}%
}
\vspace{-0.1in}
\caption{
\label{tab:additional_ablation}
\textbf{Task-Ablation Results}: We train \model\ on different combinations of tasks on \cocogpv\ split for better understanding of generalization of concepts across tasks. *Concepts seen during training refers to concepts that the model has seen in \emph{any} of the tasks that it was trained on. \best{Best} and \sbest{second best} numbers are highlighted.}
\vspace{-0.1in}

\end{table*}

\begin{figure}[t]
\centering
\includegraphics[width=0.7\linewidth]{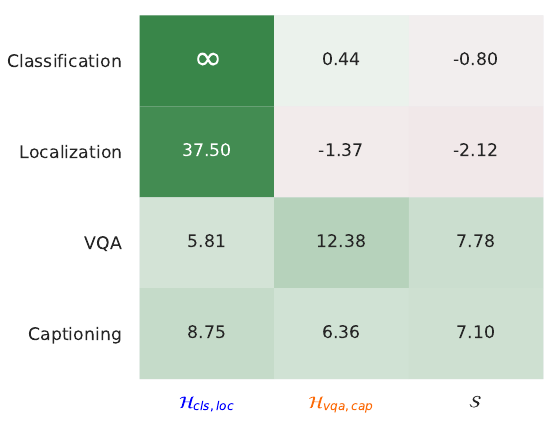}
\caption{
\label{fig:task_heatmap}
{\bf Impact of Multi-task training:} The heatmap shows the relative gain in average performance of \model-multi-task over \model-single-task for each task and category group. The average performance for any group is computed by averaging performance computed for each category in the group. 
}
\end{figure}

Fig.~\ref{fig:task_heatmap} shows the gains from multi-task training over single task performance for each task and concept-group. Multi-task training improves performance on VQA and Captioning for all groups, improves Classification performance on \hclsloc\ without impacting other groups, and improves Localization performance on \hclsloc\ at a slight cost to other groups.

We also experimented with removing the relevance-conditioning (Tab.~\ref{tab:rel_ablation}) and found the performance to be slightly better without it. This requires further exploration as conceptually we would like relevance scores to be incorporated in text prediction and for text supervision to guide the learning of region-relevance when box supervision is unavailable.

In the vision-language literature, it is common to use task-specific output heads. Tab.~\ref{tab:single_vs_multi_head} compares modality-specific output heads with task-specific heads without changing the rest of the \model\ architecture. Besides making the architecture more general-purpose, these results show that using modality-specific heads does not sacrifice performance on seen concepts but improves performance on held-out concepts by enabling transfer across tasks.   

\begin{table}[t]
\setlength{\tabcolsep}{5pt}
\small
\centering
\resizebox{0.9\columnwidth}{!}{%
\begin{tabular}{l>{\columncolor{ColorVQA}}c>{\columncolor{ColorCap}}c>{\columncolor{ColorLoc}}c>{\columncolor{ColorCls}}c}
\toprule
 & VQA & Cap. & Loc. & Class. \\
\midrule
\textcolor{tabindex}{[a]} Multitask \model\        & 58.8                        & 0.908                       & 64.7                                  & 75.4                                    \\
\textcolor{tabindex}{[b]} \hspace{0.4cm} \emph{w/o rel. cond.}     & 59.2                                 & 0.926                                 & 65.0                       & 75.9                           \\
\bottomrule
\end{tabular}%
}
\vspace{-0.05in}
\caption{
\label{tab:rel_ablation}
\textbf{Relevance conditioning ablation.} Counter-intuitively, relevance conditioning slightly hurts the performance. Further exploration is needed to utilize region-relevance scores in text prediction and to learn relevance from text supervision. 
}
\vspace{-0.2in}
\end{table}

\begin{table*}[t]
\setlength{\tabcolsep}{3pt}
\small
\centering
\resizebox{\linewidth}{!}{
\begin{tabular}{ll>{\columncolor{ColorVQA}}c>{\columncolor{ColorVQA}}c>{\columncolor{ColorVQA}}c>{\columncolor{ColorCap}}c>{\columncolor{ColorCap}}c>{\columncolor{ColorCap}}c>{\columncolor{ColorLoc}}c>{\columncolor{ColorLoc}}c>{\columncolor{ColorLoc}}c>{\columncolor{ColorCls}}c>{\columncolor{ColorCls}}c>{\columncolor{ColorCls}}c}

\toprule
\multicolumn{1}{l}{} & \multicolumn{1}{l}{} & \multicolumn{3}{c}{\cellcolor[HTML]{E6B8AF}VQA}                                       & \multicolumn{3}{c}{\cellcolor[HTML]{FFF2CC}Captioning}                                               & \multicolumn{3}{c}{\cellcolor[HTML]{CFE2F3}Localization}                          & \multicolumn{3}{c}{\cellcolor[HTML]{D9D2E9}Classification}                           \\

\multicolumn{1}{l}{Model} & \multicolumn{1}{l}{Params}& \cellcolor[HTML]{E6B8AF}Test        & \cellcolor[HTML]{E6B8AF}\emph{Seen} & \cellcolor[HTML]{E6B8AF}\emph{Unseen} & \cellcolor[HTML]{FFF2CC}Test & \cellcolor[HTML]{FFF2CC}\emph{Seen} & \cellcolor[HTML]{FFF2CC}\emph{Unseen} & \cellcolor[HTML]{CFE2F3}Test & \cellcolor[HTML]{CFE2F3}\emph{Seen} & \cellcolor[HTML]{CFE2F3}\emph{Unseen} & \cellcolor[HTML]{D9D2E9}Test & \cellcolor[HTML]{D9D2E9}\emph{Seen} & \cellcolor[HTML]{D9D2E9}\emph{Unseen} \\
\midrule
\textcolor{tabindex}{[a]} Head per Task  & 311M        & 57.67 &	58.20 &	45.86 &	\textbf{0.884} &	\textbf{0.922} &	0.533 &	62.05 &	65.76 &	26.13 &	74.26 &	\textbf{81.93} &	0.00                            \\
\textcolor{tabindex}{[b]} Head per Modality & 236M         & \textbf{57.73} &	\textbf{58.22} &	\textbf{46.91} &	0.881 &	0.915 &	\textbf{0.547} &	\textbf{62.53} &	\textbf{66.13} &	\textbf{27.75} &	\textbf{74.58} &	81.76 &	\textbf{5.10}\\
\bottomrule
\end{tabular}}
\vspace{-0.1in}
\caption{
\label{tab:single_vs_multi_head}
\textbf{Modality-specific vs. task-specific heads.} Comparing GPV-1 (\textit{b}) that uses modality-specific heads to the same architecture but with task-specific heads (\textit{a}). Both models were trained to 20 epochs. Across all tasks, modality-specific heads achieve performance comparable to task-specific heads on seen concepts, while consistently performing better on unseen concepts. 
}
\vspace{-0.2in}
\end{table*}
\section{Task descriptions}\label{sec:task_descr}
\begin{table}[t]
\scriptsize
\centering
\resizebox{0.9\columnwidth}{!}{%
\begin{tabular}{ll}
\toprule
\textbf{Task}           & \textbf{Task descriptions}                                   \\ \midrule
Captioning     & Generate a caption.                                 \\
               & Generate a description.                             \\
               & Describe this image.                                \\
               & Describe the image.                                 \\
               & Caption this image.                                 \\
               & Caption the image.                                  \\
               & What is happening in this image?                    \\
               & What is happening in the image?                     \\
               & What is going on in this image?                     \\
               & What is going on in the image?                      \\
               & Generate a caption for this image.                  \\
               & Generate a caption for the image.                   \\
               & Generate a description for this image.              \\ \midrule
Classification & What is this?                                       \\
               & What is this object?                                \\
               & What object is this?                                \\
               & What is this thing?                                 \\ \midrule
Localization   & Locate {[}OBJECT{]}.                                \\
               & Locate {[}OBJECT{]} in the image.                   \\
               & Locate {[}OBJECT{]} in this image.                  \\
               & Locate instances of {[}OBJECT{]}.                   \\
               & Locate instances of {[}OBJECT{]} in the image.      \\
               & Locate instances of {[}OBJECT{]} in this image.     \\
               & Locate all instances of {[}OBJECT{]}.               \\
               & Locate all instances of {[}OBJECT{]} in the image.  \\
               & Locate all instances of {[}OBJECT{]} in this image. \\
               & Find {[}OBJECT{]}.                                  \\
               & Find {[}OBJECT{]} in the image.                     \\
               & Find {[}OBJECT{]} in this image.                    \\
               & Find instances of {[}OBJECT{]}.                     \\
               & Find instances of {[}OBJECT{]} in the image.        \\
               & Find instances of {[}OBJECT{]} in this image.       \\
               & Find all instances of {[}OBJECT{]}.                 \\
               & Find all instances of {[}OBJECT{]} in the image.    \\
               & Find all instances of {[}OBJECT{]} in this image.   \\ \midrule
VQA            & Questions                                           \\ \midrule
RefExp         & Referring expressions                               \\ \bottomrule
\end{tabular}%
}
\caption{\textbf{Task Descriptions.} For localization prompts \mbox{[OBJECT]} is replaced with the object category name to localize. }
\vspace{-1em}
\label{tab:task_descr}
\end{table}
Tab.~\ref{tab:task_descr} lists the tasks descriptions used to create samples from annotations for each of the 5 tasks. Training on more natural, diverse, and complex task descriptions involving a wide range of skills could lead to improved ability to understand novel descriptions and better zero-shot transfer performance.  

\section{Dataset statistics}\label{sec:stats}
The sizes of \coco\ and \cocogpv\ data splits is shown in Tab.~\ref{tab:stats}. 
\begin{table}[]
\footnotesize
\setlength{\tabcolsep}{4pt}
\centering
\resizebox{1\columnwidth}{!}{%
\begin{tabular}{ll
>{\columncolor[HTML]{EFEFEF}}c 
>{\columncolor[HTML]{EFEFEF}}c 
>{\columncolor[HTML]{C0C0C0}}c 
>{\columncolor[HTML]{C0C0C0}}c 
>{\columncolor[HTML]{C0C0C0}}c }
\toprule
                             & \multicolumn{1}{l}{} & \multicolumn{2}{c}{\cellcolor[HTML]{EFEFEF}\textbf{Train}} & \multicolumn{3}{c}{\cellcolor[HTML]{C0C0C0}\textbf{Val}}                                              \\\midrule
\multicolumn{1}{l}{\coco}     & VQA                  & \multicolumn{2}{c}{\cellcolor[HTML]{EFEFEF}443757}         & \multicolumn{3}{c}{\cellcolor[HTML]{C0C0C0}214354}                                                    \\ 
                             & Captioning           & \multicolumn{2}{c}{\cellcolor[HTML]{EFEFEF}414113}         & \multicolumn{3}{c}{\cellcolor[HTML]{C0C0C0}202654}                                                    \\
                             & Localization         & \multicolumn{2}{c}{\cellcolor[HTML]{EFEFEF}241035}         & \multicolumn{3}{c}{\cellcolor[HTML]{C0C0C0}116592}                                                    \\
                             & Classification       & \multicolumn{2}{c}{\cellcolor[HTML]{EFEFEF}241035}         & \multicolumn{3}{c}{\cellcolor[HTML]{C0C0C0}116592}                                                    \\ \midrule
                             & \multicolumn{1}{l}{} & \textbf{Train}                & \textbf{Val}               & \textbf{Test} & \textbf{Seen} & \multicolumn{1}{l}{\cellcolor[HTML]{C0C0C0}\textbf{Unseen}} \\ \midrule
\multicolumn{1}{l}{\cocogpv} & VQA                  & 339411                        & 85858                      & 214354        & 205138             & 9216                                                             \\
                             & Captioning           & 294028                        & 73773                      & 202654        & 179402             & 23252                                                            \\
                             & Localization         & 174538                        & 44283                      & 116592        & 105668             & 10924                                                            \\
                             & Classification       & 174538                        & 44283                      & 116592        & 105668             & 10924                                                            \\ \bottomrule 
\end{tabular}%
}
\caption{\textbf{Dataset sizes.} The number of examples in each data split of \coco\ and \cocogpv\ for each task. Test combines Test Seen and Test Unseen.}
\label{tab:stats}
\end{table}
The division of the 80 \coco\ classes into \cocogpv\ splits (\hvqacap, \hclsloc\ and \shared) is shown in Tab.~\ref{tab:categories}.
\begin{table}[t]
\footnotesize
\centering
\setlength{\tabcolsep}{3pt}
\resizebox{1\columnwidth}{!}{%
\begin{tabular}{ll|llll}
\toprule
\textbf{Set}    & \textbf{Categories}   & \textbf{Set}  & & \textbf{Categories} & \\ \midrule
\hvqacap     & bed  & \shared   & airplane & dog & sandwich \\
               & bench  & & apple & elephant & scissors \\
               & book   & & backpack & fire hydrant & sink \\
               & cell phone & & baseball glove & fork & skateboard \\
               & horse  &   & bear & frisbee  & skis \\
               & remote &   & bicycle  & giraffe & snowboard \\
               & sheep  &   & bird & hair drier & spoon \\
               & suitcase   &   & boat  & handbag & sports ball \\
               & surfboard  &   & bowl & kite & stop sign \\
               & wine glass &  & bus  & knife & teddy bear \\
               \cline{1-2}
\hclsloc & banana       & & cake   & microwave & tennis racket \\
               & baseball bat   &  & car & motorcycle & tie \\
               & bottle &  & carrot & mouse & toaster \\
               & broccoli & & cat   & orange & toilet \\ 
               & donut  & & chair   & oven & toothbrush \\
               & hot dog &    & clock & parking meter & traffic light \\ 
               & keyboard & & couch & person & truck \\ 
               & laptop   &  & cow & pizza  & umbrella \\ 
               & train    &  & cup  & potted plant & vase \\ 
               & tv       &   & dining table & refrigerator & zebra \\ 
               \bottomrule
\end{tabular}%
}
\caption{\textbf{\cocogpv\ splits.} The 3 disjoint sets that the 80 classes of \coco\ are split into: \hvqacap\ (held-out from the VQA and captioning tasks in the  train/val sets), \hclsloc\ (held-out from the classification and localization tasks in the train/val sets) and \shared\ (not held out from any tasks).}
\label{tab:categories}
\end{table}

\section{Potential Negative Impact}\label{sec:neg_impact}
A general purpose system aims to solve the same set of tasks with a single architecture that the AI community is creating or has already created separate specialized models for. Hence, any general purpose system inherits the ethical and moral challenges faced by any special purpose system it seeks to replace. For instance, vision-language models are known to reflect~\cite{Dancette2021BeyondQB} and even amplify~\cite{Zhao2017MenAL} biases in datasets. In addition, as GPVs become increasingly capable and lower the barrier for training and using deep learning systems, a much larger portion of the population may have access to powerful data-driven capabilities. While increased accessibility may empower many without specialized training in AI or even computer science to benefit from AI tools, this democratization may make regulation of fair and ethical use of such systems challenging. Finally, high computational requirements, which in turn lead to greater energy consumption and carbon emissions\cite{Schwartz2020GreenAI}, need to be kept in check for the health of our planet and its climate.
\end{document}